# For Solving Linear Equations Recombination is a Needless Operation in Time-Variant Adaptive Hybrid Algorithms

## Abstract

Recently hybrid evolutionary computation (EC) techniques are successfully implemented for solving large sets of linear equations. All the recently developed hybrid evolutionary algorithms, for solving linear equations, contain both the recombination and the mutation operations. In this paper, two modified hybrid evolutionary algorithms contained time-variant adaptive evolutionary technique are proposed for solving linear equations in which recombination operation is absent. The effectiveness of the recombination operator has been studied for the time-variant adaptive hybrid algorithms for solving large set of linear equations. Several experiments have been carried out using both the proposed modified hybrid evolutionary algorithms (in which the recombination operation is absent) and corresponding existing hybrid algorithms (in which the recombination operation is present) to solve large set of linear equations. It is found that the number of generations required by the existing hybrid algorithms (i.e. the Gauss-Seidel-SR based time variant adaptive (GSBTVA) hybrid algorithm and the Jacobi-SR based time variant adaptive (JBTVA) hybrid algorithm) and modified hybrid algorithms (i.e. the modified Gauss-Seidel-SR based time variant adaptive (MGSBTVA) hybrid algorithm and the modified Jacobi-SR based time variant adaptive (MJBTVA) hybrid algorithm) are comparable. Also the proposed modified algorithms require less amount of computational time in comparison to the corresponding existing hybrid algorithms. As the proposed modified hybrid algorithms do not contain recombination operation, so they require less computational effort, and also they are more efficient, effective and easy to implement.

**Major area of the paper:** Algorithm

**Statement of Achievement:** The time variant adaptive Hybrid algorithms require no Recombination operation for solving linear equations.

## 1. Introduction

Solving a large set of simultaneous linear equations is a fundamental problem that occurs in diverse applications in engineering and science, as well as with applications of mathematics to the social sciences and the quantitative study of business, statistical and economic problems. Also for appropriate decision making related to the same physical problem, sometimes of the physical an appropriate algorithm which converges rapidly and efficiently for solving large set of linear equations is desired. For example, in cases of short-term weather forecast, image processing, simulation to predict aerodynamics performance where solution of very large set of simultaneous linear equations by numerical methods are required and time is an important factor for practical application of the results. For large set of linear equations, especially for sparse and structured coefficients, iterative methods are preferable, as iterative method are unaffected by round off errors [1]. The rate of convergence of the well-known classical numerical iterative methods namely the Jacobi method and the Gauss-Seidel method is very slow for the both methods and can be accelerated by using successive relaxation (SR) technique [2, 3]. But the speed of convergence depends on the relaxation factor $\omega$ $(0 < \omega < 2)$ and SR technique is very much sensitive to the relaxation factor [4, 5]. However, it is often very difficult to estimate the optimal relaxation factor, which is a key parameter of the SR technique [4, 6].

On the other hand the evolutionary algorithms (EA) are developed from some natural phenomena: genetic inheritance and Darwinian strife for survival [7, 8, 9]. Obvious biological evidence is that a rapid change is observed at early stages of life and a slow change is observed at later stages of life in all kinds of animals/plants. These changes are more often occurred dynamically depending on the situation exposed to them. By mimicking this emergent natural evidence, a special dynamic time-variant mutation (TVM) operator is proposed by Hashem [10], Watanabe and Hashem [11] and Michalewicz et. al. [12,13, 14] in global optimization problems. Generally, most of the works on EA can be classified as evolutionary optimization (either numerical or combinatorial) or evolutionary learning.

Recently, Gauss-Seidel based uniform adaptive (GSBUA) hybrid evolutionary algorithm [15] and Jacobi based uniform adaptive (JBUA) hybrid evolutionary algorithm [16] have been developed for solving large set of linear equations by integrating classical numerical methods with uniform adaptive (UA) evolutionary techniques. In these algorithms both recombination and mutations operations are present. Furthermore, Gauss-Seidel based Time variant adaptive (GSBTVA) hybrid evolutionary algorithm [17] and Jacobi based Time variant adaptive (JBTVA) hybrid evolutionary algorithm [18] have been developed for solving large set of linear equations by integrating classical numerical methods with Time variant adaptive (TVA) evolutionary techniques. In these algorithms both recombination and mutations operations are also present. The idea of self-adaptation was also applied in many different fields [14, 19, 20]. Moreover Fogel and Atmar [21] used linear equation solving as test problems for comparing recombination, inversion operations and Gaussian mutation in an evolutionary algorithm. However, the emphasis of their study was not on equation solving, but rather on comparing the effectiveness of recombination relative to mutation and only small size of problems ($n = 10$) was considered [21].

In this paper, two modified time variant hybrid evolutionary algorithms (i.e. modified GSBTVA and the modified JBTVA algorithms) are proposed to solve large set of linear equations. The proposed modified hybrid algorithms are modified from the existing GSBTVA and the JBTVA algorithms and contain all the evolutionary operations valuable in the existing algorithms except recombination operation. The proposed modified hybrid algorithms initialize uniform relaxation factors in a given domain and "evolve" it by time-variant adaptation technique as well. The main mechanisms of the proposed modified algorithms are initialization, mutation, time variant adaptation, and selection mechanisms (i.e. recombination operation is absent). It makes better use of a population by employing different equation-solving strategies for different individuals in the population. The errors are minimized by mutation and selection mechanisms.

We have tried to investigate the necessity of the recombination operation presented in the existing time variant adaptive hybrid evolutionary algorithms [17, 18] and also to investigate where absent of recombination operations hamper the process. The effectiveness of the proposed modified hybrid algorithms is compared with that of corresponding the existing GSBTVA and the JBTVA hybrid evolutionary algorithms. The preliminary investigation has showed that the both proposed modified algorithms are comparable with the corresponding existing hybrid evolutionary algorithms in terms of number of generation required for expected result. Also the proposed modified algorithms are more efficient and effective than the corresponding existing GSBTVA and JBTVA algorithms in terms of computational effort. Moreover the proposed modified hybrid algorithms are more easy to implement and required less memory allocation.

## 2. The Basic Equations of Classical Methods

The system of $n$ linear equations with $n$ unknown $x_1$, $x_2$, $\cdots$,..., $x_n$ can be written as

$$\sum_{j=1}^{n} a_{ij} x_j = b_i, \; \left( i = 1, 2, \cdots, n \right)$$

or equivalently, in matrix form

$$\mathbf{Ax} = \mathbf{b} \tag{1}$$

where $\mathbf{A} \in \Re^n \times \Re^n$, $\mathbf{x} \in \Re^n$ and $\mathbf{b} \in \Re^n$, here $\Re$ is real number field.

We know that for unique solution $| \mathbf{A} | \neq 0$. Let us assume, without loss of generality, that none of the diagonal entries of $\mathbf{A}$ are zero; otherwise rows will be interchanged.

Now the coefficient matrix $\mathbf{A}$ can be decomposed as $\mathbf{A} = (\mathbf{D} + \mathbf{U} + \mathbf{L})$

where $\mathbf{D} = (d_{ij})$ is a diagonal matrix, $\mathbf{L} = (l_{ij})$ is a strictly lower triangular matrix and $\mathbf{U} = (u_{ij})$ is a strictly upper triangular matrix. Then according to classical Jacobi method, Eq.(1) can be rewritten as

$$\mathbf{x}^{(k+1)} = \mathbf{H}_j \mathbf{x}^{(k)} + \mathbf{V}_j \quad \text{with} \quad \mathbf{x}^{(k)} = \left( x_1^{(k)}, x_2^{(k)}, \cdots, x_n^{(k)} \right)^{\mathrm{t}} \quad \text{and} \quad k = 0, 1, 2, \cdots \tag{2}$$

where $\mathbf{H}_j = \mathbf{D}^{-1}(-\mathbf{L} - \mathbf{U})$ is the Jacobi iteration matrix and $\mathbf{V}_j = \mathbf{D}^{-1}\mathbf{b}$.

On introducing SR technique [3, 6] Eq. (2),can be written as

$$\mathbf{x}^{(k+1)} = \mathbf{H}_{j(\omega)} \mathbf{x}^{(k)} + \mathbf{V}_{j(\omega)} \tag{3}$$

where, $\mathbf{H}_{j(\omega)} = \mathbf{D}^{-1}\{(1-\omega)\mathbf{I} - \omega\,(\mathbf{L} + \mathbf{U})\}$, $\mathbf{V}_{j(\omega)} = \omega\,\mathbf{D}^{-1}\mathbf{b}$, , $\omega \in (\,\omega_L, \omega_U\,)$ is the relaxation factor which influences the convergence rate greatly and $\mathbf{I}$ is the identity matrix; also $\omega_L$ and $\omega_U$ are the lower and upper bound of $\omega$.

Similarly according to the classical Gauss-Seidel method by introducing SR technique [3, 6], Eq. (2), can be again written as

$$\mathbf{x}^{(k+1)} = \mathbf{H}_{g(\omega)} \mathbf{x}^{(k)} + \mathbf{V}_{g(\omega)} \tag{4}$$

where $\mathbf{H}_{g(\omega)} = (\mathbf{I} + \omega\,\mathbf{D}^{-1}\mathbf{L})^{-1}\{(1-\omega)\mathbf{I} - \omega\,\mathbf{D}^{-1}\mathbf{U}\}$ is the Gauss-Seidel-SR iteration matrix and $\mathbf{V}_{g(\omega)} = \omega\,(\mathbf{I} + \omega\,\mathbf{D}^{-1}\mathbf{L})^{-1}\mathbf{D}^{-1}\mathbf{b}$.

### 3. Time Variant Adaptive Hybrid Evolutionary Algorithms

#### 3.1 The Existing Hybrid Evolutionary Algorithms

The main aim of the hybridization of the classical SR methods with the evolutionary computation techniques is to self-adapt the relaxation factor used in the classical SR technique. And the time variant adaptation technique is used to escape from disadvantage of the uniform adaptation and for introducing fine-tuning to the rate of convergence. The relaxation factors are adapted on the basis of the fitness of individuals (i.e. how well an individual solves the equations). Similar to many other evolutionary algorithms, the hybrid algorithm always maintains a population of approximate solutions to the linear equations. Each solution is represented by an individual. The initial population is generated randomly from the field $\Re^n$. Different individuals use different relaxation factors. Recombination in the hybrid algorithm involves all individuals in a population. If the population is of size $N$, then the recombination will have $N$ parents and generates $N$ offspring through linear combination. Mutation is achieved by performing one iteration of classical (Gauss-Seidel or Jacobi) method with SR technique. The mutation is stochastic since $\omega$, used in the iteration are initially determined between $\omega_L$ (=0) and $\omega_U$ (=2), is adapted stochastically in each generation and adaptation nature is also time-variant. The fitness of an individual is evaluated on the basis of the error of an approximate solution. For example, given an approximate solution (i.e., an individual) $\mathbf{z}$, its error is defined by $\|e(\mathbf{z})\| = \|\mathbf{A}\mathbf{z} - \mathbf{b}\|$ . The relaxation factors are adapted after each generation, depending on how well an individual performs (in terms of the error). The main steps of the existing JBTVA and the GSBTVA hybrid algorithms are – Initialization, Recombination, Mutation, Adaptation and Selection mechanism [17, 18]. The pseudo-code structures of the existing hybrid evolutionary algorithms [17, 18] is given bellow:

Algorithm_JBTVA/GSBTVA()

**begin**

$t \leftarrow 0$  ; /* *Initialize the generation counter* */

Initialize population: $\mathbf{X}^{(0)} = \{\mathbf{x}_1^{(0)}, \mathbf{x}_2^{(0)} \ldots \mathbf{x}_N^{(0)}\}$ ;

/* *Here* $\mathbf{x}_i^{(t)} \Longrightarrow$ *i-th individual at t-th generation* */

Initialize relaxation factors: $\omega_i = \begin{cases} \omega_L + \dfrac{d}{2} & \text{for } i = 1 \\ \omega_{i-1} + d & \text{for } 1 < i \le N \end{cases}$ , $d = \dfrac{\omega_U - \omega_L}{N}$

Evaluate population: $||e(\mathbf{X})|| = \{||e(\mathbf{z})|| : \mathbf{z} \in \mathbf{X}\}$ ;

**While (not** termination-condition**) do**

**begin**

Select individuals for reproduction:

Apply operators:

Crossover: $\mathbf{X}^{(k+c)} = \mathbf{R}\left(\mathbf{X}^{(k)}\right)'$ ;

/* $\mathbf{R}$ is stochastic matrix & Superscript c indicates Crossover */

Mutation: $\mathbf{x}_i^{(k+m)} = \mathbf{H}_{q(\omega_i)}\mathbf{x}_i^{(k+c)} + \mathbf{V}_{q(\omega_i)}$ ;

/* where $q \in \{j,g\}$, "j" indicate Jacobi based method and "g" indicate Gauss-Seidel based method */

Evaluate newborn offspring: $||e(\mathbf{X}^{(k+m)})|| = \{||e(\tilde{\mathbf{x}}^{(k+m)})|| : \tilde{\mathbf{x}}^{(k+m)} \in \mathbf{X}^{(k+m)}\}$ ;

Adaptation of $\omega$ : $\omega_x = f_x(\omega_x,\omega_y,p_x)$ & $\omega_y = f_y(\omega_x,\omega_y,p_y)$ ;

/* $p_x$ and $p_y$ are adaptive probability functions */

Selection and reproduction: $\mathbf{X}^{(k+1)} = \varsigma(\mathbf{X}^{(k+m)})$ ;

$k \leftarrow k+1$ ; /* Increase the generation counter */

**end**

**end**

As the adaptation and the selection are the main characteristic mechanisms of the existing hybrid algorithms(as well as proposed modified algorithm also), so we have described them in brief bellow:

**Adaptation:**

Let $\mathbf{x}^{(k+m)}$ and $\mathbf{y}^{(k+m)}$ be two offspring individuals with relaxation factors $\omega_x$ and $\omega_y$ and with errors (fitness values) $\| e(\mathbf{x}^m) \|$ and $\| e(\mathbf{y}^m) \|$ respectively. Then the relaxation factors $\omega_x$ and $\omega_y$ are adapted as follows:

(a) If $\| e(\mathbf{x}^m) \| > \| e(\mathbf{y}^m) \|$,

(i) then $\omega_x$ is moved toward $\omega_y$ by setting

$$\omega_x^m = (0.5 + p_x)(\omega_x + \omega_y) \qquad (5)$$

and   (ii) $\omega_y$ is moved away from $\omega_x$ by setting

$$\omega_y^m = \begin{cases} \omega_y + p_y(\ \omega_U - \omega_y), & \text{when } \omega_y > \omega_x \\ \omega_y + p_y(\ \omega_L - \omega_y), & \text{when } \omega_y < \omega_x \end{cases} \qquad (6)$$

where $p_x = E_x \times N(0,0.25) \times T_\omega$   and   $p_y = E_y \times |N(0,0.25)| \times T_\omega$, are the time-variant adaptive (TVA) probability parameters of $\omega_x$ and $\omega_y$ respectively.

Here $\quad T_\omega = \lambda \ln(1 + \dfrac{1}{t+\lambda})$ , $\lambda > 10$ \hfill (7)

which gives the basic time-variant parameter (BTV) [18] in which $\lambda$ is an exogenous parameter. $\lambda$ is used to increase or decrease the rate of change of curvature with respect to the number of iterations $t$ (see Figure 1). Also $N(0, 0.25)$ is the Gaussian distribution with mean 0.0 and standard deviation 0.25. Now $E_x$ and $E_y$ indicate the approximate initial boundary of the variation of TVA parameters of $\omega_x$ i.e. (-$E_x$, $E_x$) and $\omega_y$ i.e. (0, $E_y$) respectively. Note that $\omega_x^m$ and $\omega_y^m$ are adapted relaxation factors correspond to $\omega_x$ and $\omega_y$ respectively.

(b) If $\| \mathrm{e}(\mathbf{x}^m) \| < \| \mathrm{e}(\mathbf{y}^m) \|$, then $\omega_x$ and $\omega_y$ are adapted in the same way as above but in the reverse order of $\omega_x^m$ and $\omega_{\mathbf{y}}^m$.

(c) If $\| \mathrm{e}(\mathbf{x}^m) \| = \| \mathrm{e}(\mathbf{y}^m) \|$, no adaptation will take place i.e.

$\omega_x^m = \ \omega_x$ and $\omega_y^m = \omega_y$.

The characteristics of BTV parameter including TVA probability parameter $p_x$ and $p_y$ are shown in Figures 1, 2 and 3 respectively.

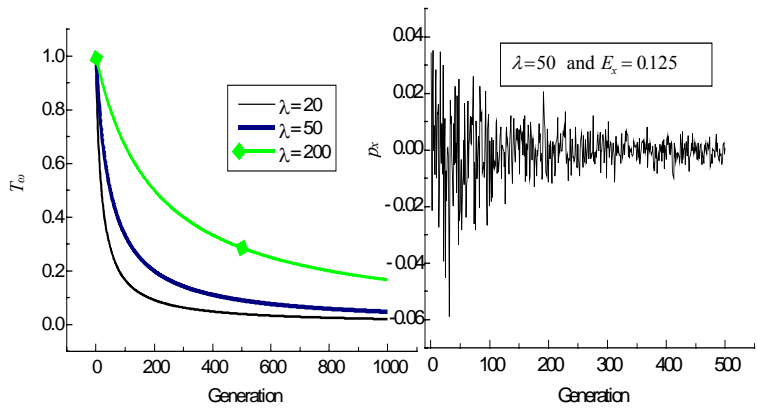
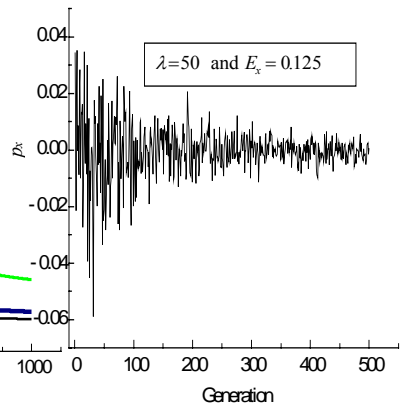
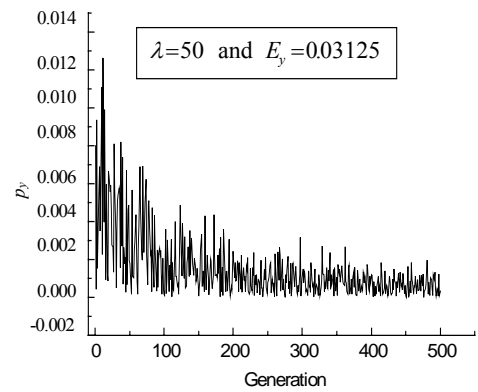

**Figure 1**        **Figure 2**        **Figure 3**

**Selection and Reproduction:**

The best $N/2$ offspring individuals are selected according to their fitness values (errors). Then the selected offspring are reproduced (i.e. each parent individuals generate two offspring). Thus the next generation of $N$ individuals is formed.

## 3.2    The Proposed Modified Hybrid Evolutionary Algorithms

The key idea behind the proposition of the modified algorithms (Modified Jacobi Based Time Variant Adaptive (MJBTVA) hybrid evolutionary algorithm and the Modified Gauss-Seidel Based Time Variant Adaptive (MGSBTVA) hybrid evolutionary algorithm is to examine the necessity of recombination operation for solving linear equations. So the proposed modified hybrid evolutionary algorithms contains all steps of the JBTVA and GSBTVA hybrid evolutionary algorithms except the step − recombination. And we do not repeat the pseudo-code structures of the both modified time variant hybrid evolutionary algorithms here.

# 4.  Performance of The Modified Algorithms

In order to evaluate the effectiveness of the proposed MJBTVA and the MGSBTVA hybrid algorithm, a number of numerical experiments have been carried out to solve the Eq. (1). The following settings were valid for all the experiments:

The dimension of unknown variable $n = 200$, Population size $N = 2$, boundary of relaxation factors $(\omega_L, \omega_U) = (0, 2)$, the approximate initial boundaries, $E_x$ and $E_y$ were set at 0.125 and 0.03125 respectively, the exogenous parameter $\lambda$ was set at 50, each individual $\mathbf{x}$ of population $\mathbf{X}$ was initialized from the domain $\Re^{200} \in (-30, 30)$ randomly and uniformly, the threshold error was $\eta$ set at $10^{-7}$ and the stochastic matrix $\mathbf{R}$ was generated randomly.

The first problem was set by considering $a_{ii} \in (100, 200)$; $a_{ij} \in (-10, 10)$; $b_i \in (100, 200)$, $i, j = 1, \cdots, n$ and (problem $P_1$ in Table 1 & 2). A single set of parameters was generated randomly from the above-mentioned problem and the following two experiments were carried out. The problem was solved with an error smaller than $10^{-7}$ (threshold error).

In the first experiment, the comparison between the JBTVA [17] and the proposed MJBTVA had been made. Figure 4 shows the numerical results of this experiment. From this experiment, two important observations came out. Firstly the proposed MJBTVA algorithm is comparable with the JBTVA algorithm in terms of generation. Secondly the proposed MJBTVA algorithm required less amount of time than that of JBTVA algorithm.

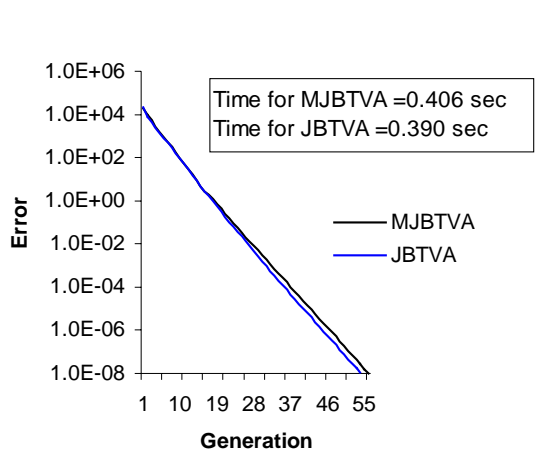

**Figure 4:** Comparison between t MJBTVA and JBTVA algorithms

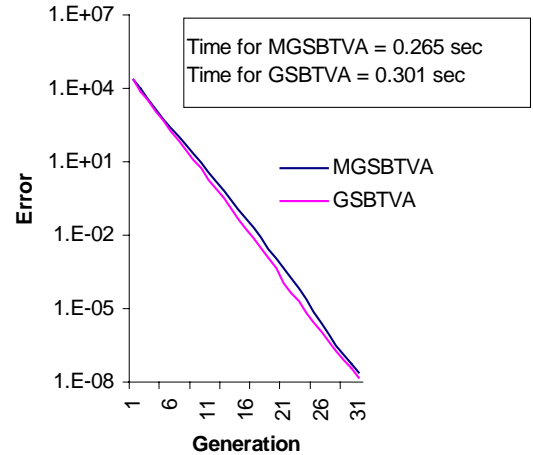

**Figure 5:** Comparison between MGSBTVA and GSBTVA algorithms

In the second experiment, the comparison between the GSBTVA [18] and the proposed MGSBTVA had been made. Figure 5 shows the numerical results of this experiment. Again two observations came out – (i) the proposed MGSBTVA algorithm is comparable with the GSBTVA algorithm in terms of generation and (ii) the proposed MGSBTVA algorithm required less amount of time than that of GSBTVA algorithm.

**Table 1**: Comparison between the JBTVA and the MJBTVA hybrid algorithms for several randomly generated test problems

| Label of Test Problems | Domain of the elements of the coefficient matrix **A** & the right side constant vector **b** of the test Problems | | | MJBTVA Alg. | | JBTVA Alg. | |
|---|---|---|---|---|---|---|---|
| | | | | Generation (Elapsed) | Elapse time (mili sec) | Generation (elapsed) | Elapse time (mili sec) |
| $P_1$ | $a_{ii} \in (100,200)$; | $a_{ii} \in (-10,10)$; | $b_i \in (100,200)$ | 57 | 390 | 58 | 406 |
| $P_2$ | $a_{ii} \in (1,400)$; | $a_{ii} \in (-4,4)$; | $b_i = 100$ | 171 | 968 | 168 | 1046 |
| $P_3$ | $a_{ii} \in (-50,50)$; | $a_{ii} \in (-1,1)$; | $b_i \in (-1,1)$ | 39 | 579 | 37 | 625 |
| $P_4$ | $a_{ii} = 100$; | $a_{ii} \in (-1,1)$; | $b_i \in (-00,100)$ | 42 | 539 | 40 | 562 |
| $P_5$ | $a_{ii} = 50$; | $a_{ii} \in (-10,10)$; | $b_i \in (-5,5)$ | 10 | 157 | 10 | 187 |
| $P_6$ | $a_{ii} = 50$; | $a_{ii} = (-1,1)$; | $b_i = 2$ | 10 | 172 | 10 | 187 |
| $P_7$ | $a_{ii} = 20i$; | $a_{ii} = (100\text{-}j) /20$ ; $b_i = 10 \, i$ | | 71 | 1125 | 84 | 1359 |
| $P_8$ | $a_{ii} = 20n$; | $a_{ii} = j$; | $b_i = i$ | 64 | 1060 | 70 | 1410 |
| $P_9$ | $a_{ii} = (-20, 200)$; | $a_{ii} \in (-2,3)$; | $b_i \in (-2,3)$ | 475 | 7422 | 489 | 7625 |
| $P_{10}$ | $a_{ii} = 40$; | $a_{ii} \in (-4,4)$; | $b_i = 200$ | 74 | 1203 | 75 | 1250 |
| $P_{11}$ | $a_{ii} \in (-50,50)$; | $a_{ii} \in (-1,1)$; | $b_i \in (-1,1)$ | 35 | 531 | 38 | 625 |

**Table 2**: Comparison between the GSBTVA and the proposed MGSBTVA hybrid
algorithms for several randomly generated test problems

| Label of Test Problems | Domain of the elements of the coefficient matrix **A** & the right side constant vector **b** of the test Problems | | | MGSBTVA Alg. | | GSBTVA Alg. | |
|---|---|---|---|---|---|---|---|
| | | | | Generation (Elapsed) | Elapse time (mili sec) | Generation (elapsed) | Elapse time (mili sec) |
| $P_1$ | $a_{ii} \in (100,200)$; | $a_{ij} \in (-10,10)$; | $b_i \in (100,200)$ | 31 | 265 | 32 | 301 |
| $P_2$ | $a_{ii} \in (1,400)$; | $a_{ij} \in (-4,4)$; | $b_i = 100$ | 105 | 828 | 108 | 843 |
| $P_3$ | $a_{ii} \in (-50,50)$; | $a_{ij} \in (-1,1)$; | $b_i \in (-1,1)$ | 39 | 562 | 42 | 579 |
| $P_4$ | $a_{ii} = 100$; | $a_{ij} \in (-1,1)$; | $b_i \in (-00,100)$ | 39 | 485 | 41 | 578 |
| $P_5$ | $a_{ii} = 50$; | $a_{ij} \in (-10,10)$; | $b_i \in (-5,5)$ | 08 | 141 | 08 | 156 |
| $P_6$ | $a_{ii} = 50$; | $a_{ij} \in (-1,1)$; | $b_i = 2$ | 13 | 203 | 13 | 219 |
| $P_7$ | $a_{ii} = 20i$; | $a_{ij} = (100\text{-}j)$ /20 ; $b_i = 10$ i | 117 | 1500 | 116 | 1547 |
| $P_8$ | $a_{ii} = 20n$; | $a_{ij} = j$; | $b_i = i$ | 64 | 875 | 69 | 968 |
| $P_9$ | $a_{ii} = (-20, 200)$; | $a_{ij} \in (-2,3)$; | $b_i \in (-2,3)$ | 571 | 736 | 583 | 7547 |
| $P_{10}$ | $a_{ii} = 40$; | $a_{ij} \in (-4,4)$; | $b_i = 200$ | 109 | 1439 | 111 | 1469 |
| $P_{11}$ | $a_{ij} \in (-50,50)$; | $a_{ij} \in (-1,1)$; | $b_i \in (-1,1)$ | 13 | 110 | 13 | 141 |

Table 1and 2 represent eleven test problems, labeled from $P_1$ to $P_{11}$, with dimension, $n = 200$. For each test problem $P_i$: $i$ = 1, 2, . . ., 11, the coefficient matrix **A** and constant vector **b** were all generated uniformly and randomly within given domains. **Table 1** shows the comparison between the number of generation (iteration) of the JBTVA and the proposed MJBTVA hybrid algorithms to the given threshold error, $\eta$. One observation can be made immediately from this table that the proposed MJBTVA hybrid algorithm is comparable with the existing JBTVA hybrid algorithm for all the problems. Another observation is that the proposed MJBTVA required less amount of time than that of JBTVA for all the problems.

**Table 2** shows the comparison between the number of generation (iteration) of the GSBTVA and the proposed MGSBTVA hybrid algorithms to the given threshold, $\eta$. One observation can be made immediately from this table that the proposed MGSBTVA hybrid algorithm is comparable with the GSBTVA hybrid algorithm for all the problems. Another observation is that the proposed MGSBTVA required less amount of time than that of MGSBTVA for all the problems.

It is to be mentioned here that a total of ten independent runs with different sample paths were conducted. The average results are reported here. Also for all the experiments, the times were measured in the same environment.

## 5. Concluding Remarks

In this paper, two modified ( Jacobi based and Gass-Seidel based) hybrid evolutionary algorithm with time-variant adaptive (TVA) technique has been proposed for solving large set of linear equations. These proposed MJBTVA and MGSBTVA hybrid algorithms have been modified by omitted completely the recombination operation from the JBTVA and the GSBTVA hybrid algorithms respectively. The effectiveness of the proposed modified algorithms is compared with that of the JBTVA and the GSBTVA hybrid algorithms respectively. This preliminary investigation has showed that both the proposed MJBTVA and MGSBTVA hybrid algorithms are comparable in terms of generation (iteration) with the JBTVA and GSBTVA respectively. Also both the proposed MJBTVA and MGSBTVA hybrid algorithms required less amount of time than the JBTVA and GSBTVA hybrid algorithm respectively. Furthermore since proposed modified hybrid algorithms have no recombination operation, so they require less memory allocation and computational effort to solve the problems. Moreover, the proposed modified hybrid algorithms are also very simple and easier to implement both in sequential and parallel computing environments.